\documentclass[journal,twoside,web]{ieeecolor}
\usepackage{colortbl}
\usepackage{generic}
\usepackage{cite}
\usepackage{amsmath,amssymb,bm}
\usepackage{graphicx}
\usepackage{booktabs,array,multirow}
\usepackage{url}
\makeatletter
\let\NAT@parse\undefined
\makeatother
\usepackage{hyperref}
\definecolor{citationblue}{rgb}{0.00,0.34,0.93}
\definecolor{referencegreen}{rgb}{0.10,0.36,0.26}
\hypersetup{
    colorlinks=true,
    citecolor=citationblue,
    linkcolor=referencegreen,
    urlcolor=black,
    pdfauthor={Xiaokai Bai, Songkai Wang, Lianqing Zheng, Runwei Guan, Siyuan Cao, Hui-liang Shen},
    pdftitle={RC-GeoCP: Geometric Consensus for 4D Radar--Camera Collaborative Perception},
}
\newcolumntype{C}[1]{>{\centering\arraybackslash}p{#1}}

\newcommand{\softmax}{\operatorname{Softmax}}

\definecolor{apfirst}{RGB}{139,206,182}
\definecolor{apsecond}{RGB}{214,241,227}
\definecolor{apthird}{RGB}{255,246,204}
\newcommand{\apfirst}[1]{\cellcolor{apfirst}\textbf{#1}}
\newcommand{\apsecond}[1]{\cellcolor{apsecond}\underline{#1}}
\newcommand{\apthird}[1]{\cellcolor{apthird}#1}
\title{RC-GeoCP: Geometric Consensus for 4D Radar--Camera Collaborative Perception}
\author{
    Xiaokai Bai\textsuperscript{1}, Songkai Wang\textsuperscript{1}, Lianqing Zheng\textsuperscript{2}, Runwei Guan\textsuperscript{3}, Siyuan Cao\textsuperscript{1}, Hui-liang Shen\textsuperscript{1}
    \thanks{
    \textsuperscript{1}College of Information Science and Electronic Engineering, Zhejiang University. 
    \textsuperscript{2}School of Automotive Studies, Tongji University. 
    \textsuperscript{3}Thrust of Artificial Intelligence, Hong Kong University of Science and Technology.
    Contact e-mail: \texttt{shawnnnkb@gmail.com}.}
}
\begin{document}
\bstctlcite{TIIReferenceStyle}
\maketitle
\begin{abstract}
    In industrial transport, vehicles and roadside units must select which
    bird's-eye-view (BEV) features to share over bandwidth-limited links.
    Camera depth errors can misplace confident responses. Across agents,
    4D radar observations of the same object provide a common geometric
    reference for its location. In existing methods, motion-guided attention can downweight
    stationary targets, while early averaging obscures which agent provides
    spatial support. The receiver may then spend bandwidth on misplaced
    features and give them undue weight during fusion.
    To address this challenge, we propose RC-GeoCP. First, Geometric
    Structure Rectification (GSR) uses local radar to ground image features
    in BEV. Each neighbor shares foreground confidence and radar support
    maps before feature transmission. Next, Uncertainty-Aware Communication
    (UAC) combines receiver demand with confidence and radar support from
    the same neighbor at each cell. It uses the resulting geometric
    consensus score to request the highest-ranked cells on each link.
    Finally, the Consensus-Driven Assembler (CDA) reuses this score to
    weight sources at each cell during fusion. With common radar--camera
    inputs, RC-GeoCP achieves the highest BEV average precision (AP) among
    the evaluated methods on the V2X-Radar test set and the V2X-R validation
    set. On V2X-Radar, sparse exchange retains 98.6\% of dense validation
    AP@0.7 with 63.8\% fewer bytes per link, while on V2X-R, it retains 99.1\%
    with 69.3\% fewer bytes per link.
\end{abstract}

\section{Introduction}
\label{sec:introduction}

\IEEEPARstart{I}{n} connected industrial transport, roadside infrastructure
and mobile agents share perception to extend visibility beyond individual
fields of view~\cite{OpenV2XTII2026,RoadsideRepresentationTII2025}.
A roadside camera, for example, can observe traffic participants hidden
from the vehicle~\cite{OPV2V,DAIR-V2X,V2VNet,V2X-ViT}. Rain or fog can
reduce visibility from both viewpoints. Under these conditions, 4D radar
still provides direct range, elevation and radial-velocity measurements
with lower sensitivity to poor optical visibility~\cite{Kradar,RadarCP}.
These measurements complement the semantic cues from cameras when agents
share observations~\cite{appealing}. Under limited bandwidth, the receiver
selects which features to request, then decides how much each source
contributes during fusion.

Realizing this benefit depends on how accurately each agent localizes
the evidence it shares. Uncertain
image depth can place a confident object response in the wrong
bird's-eye-view (BEV) cell~\cite{LSS,BEVDepth}. When the receiver aligns a
neighbor's features to its own grid, it inherits any such placement errors.
In blind regions, the receiver lacks local observations to check this
placement. A misplaced response can then consume a transmission slot and
reinforce the wrong location during fusion.
LiDAR, widely used in collaborative perception~\cite{V2X-R}, supplies
precise depth at pixels covered by its returns. BM2CP, for example, uses
a neighbor's LiDAR measurements to fill gaps left by the local scan in the
image depth map~\cite{BM2CP}. However, deploying LiDAR on both vehicles
and roadside infrastructure increases sensing costs~\cite{LXL}.
Heavy rain or dense fog can also reduce its return coverage, depending
on target range and reflectivity~\cite{KimLiDARRainFog}.
We focus on vehicle--roadside systems that use cameras and 4D radar at
inference. Compared with LiDAR, sparse 4D radar provides fewer spatial
anchors for visual features, leaving their placement more dependent on
uncertain image depth.

\begin{figure}[t]
    \centering
    \includegraphics[width=\columnwidth]{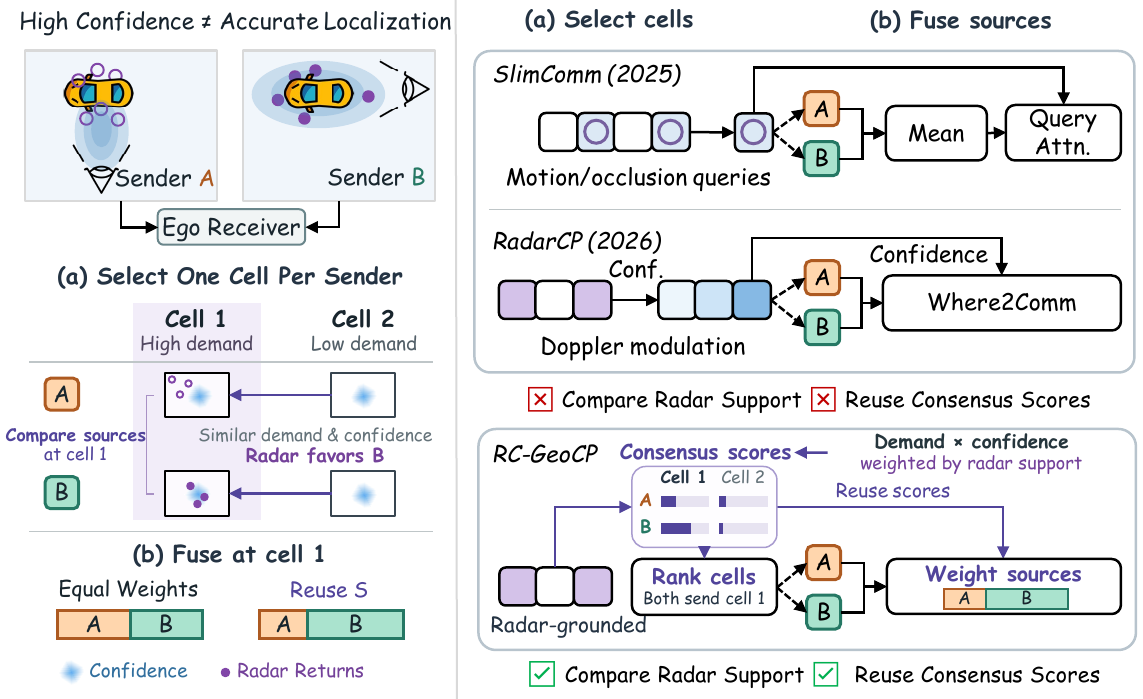}
    \caption{Each neighbor can send one BEV cell. Both send cell~1, where
    A and B report similar foreground confidence but the radar returns
    provide stronger spatial support for B. Reusing the consensus score
    gives B more weight during fusion.}
    \label{fig:head}
\end{figure}

Feature placement therefore matters both when the receiver requests cells
under a link budget and when it weights sources during fusion.
Where2comm uses spatial confidence to guide region selection and
fusion~\cite{Where2comm}, while CoSDH uses receiver-side point-cloud density
to identify regions that need evidence from confident
sources~\cite{CoSDH}. With sparse radar, a region containing a vehicle can yield as few
returns as an empty region. Point density alone then gives the receiver
little basis for deciding whether to request that region.
Radar-informed approaches use sparse returns to ground local visual
features or guide region selection~\cite{LXL,RadarCP,SlimComm}.
Local fusion alone does not tell the receiver how strongly each source's
radar observations support a confident response at a given location.

When 4D radars on different agents observe the same object, their range
and angle measurements provide a common geometric reference for the
region it occupies in BEV. This position reference also remains useful
for stationary objects, even when their radial velocity is low.
The receiver can thus assess each foreground
response against radar support from the same agent. Pooling radar support
would erase this correspondence, removing the source-specific geometric
evidence needed to distinguish similarly confident responses.
In SlimComm, attention operates on the mean response from neighbors
to each motion- or occlusion-guided query~\cite{SlimComm}. RadarCP emphasizes motion with a Doppler
mask~\cite{RadarCP}, which can downplay position cues from stationary
objects. \emph{We therefore study how a receiver can use each neighbor's radar
support to decide which features to request under a transmission budget
and how much weight to give them once they arrive.}

We address this challenge through \emph{radar-guided geometric consensus}.
Before requesting features, the receiver obtains a spatial support map
predicted from the 4D radar measurements of each neighbor. At each BEV
cell, geometric consensus captures the extent to which a foreground
prediction and radar observations from the same agent jointly support
occupancy. Because even well-supported features can duplicate local evidence,
receiver demand modulates this joint support before the receiver ranks cells within each link quota and
weights sources at each cell after delivery.
Figure~\ref{fig:head} illustrates why the receiver retains radar
support from each neighbor even when both send the same cell. Geometric Structure Rectification (GSR)
adapts radar--camera queries and depth-aware retrieval~\cite{SGDet3D}
to ground visual features before sharing. Uncertainty-Aware Communication
(UAC) then ranks the cells, and the Consensus-Driven Assembler (CDA) reuses
the score after delivery. Their shared spatial cue is learned from
receiver-frame occupancy and detection supervision. Reusing the score
preserves each source's geometric support from selection through fusion.
It also lets detection feedback train the source preference used to
request features.
Our contributions are as follows.

\begin{itemize}
    \setlength{\parfillskip}{0pt}
    \item We propose RC-GeoCP, a geometric consensus framework for
    4D radar--camera collaborative perception. It pairs each BEV foreground
    response with radar support from the same agent to guide feature
    exchange and fusion.
    \item We develop GSR to guide depth-aware image sampling with local
    4D radar. A learned gate then controls how much radar information
    enters the retrieved visual features at each BEV cell before
    transmission.
    \item We develop a shared score for UAC and CDA that combines
    receiver demand with confidence and radar support from the same agent.
    It ranks cells within each link quota and guides source weighting
    during fusion.
    \item We show detection gains across three fusion backbones on V2X-Radar and
    V2X-R. At fixed traffic, a shared score for selection and fusion
    improves validation AP over independently trained scores on both
    datasets.
\end{itemize}

\section{Related Work}
\label{sec:related}

\subsection{Local 4D Radar--Camera Fusion}

Image depth errors can displace visual features in BEV before agents
share them. Local radar measurements provide a range reference for
sampling and placing those features.

BEVFusion uses predicted depth to lift image features to a BEV grid
shared with LiDAR~\cite{BEVFusion}. RCFusion uses a common grid
for radar--camera fusion~\cite{RCFusion}, and RCBEVDet aligns the two
modalities with deformable attention~\cite{RCBEVDet}. RCGDet3D predicts
Gaussian attributes in ray-aligned coordinates and incorporates image
semantics into the radar features~\cite{RCGDet3D}. LXL guides image
sampling with radar occupancy and predicted depth~\cite{LXL}, while
SGDet3D retrieves image features with multimodal BEV queries for
single-agent detection~\cite{SGDet3D}. RC-GeoCP retains local radar
support after feature grounding to guide collaborative requests and fusion.

\begin{figure*}[t]
    \centering
    \includegraphics[width=0.95\textwidth]{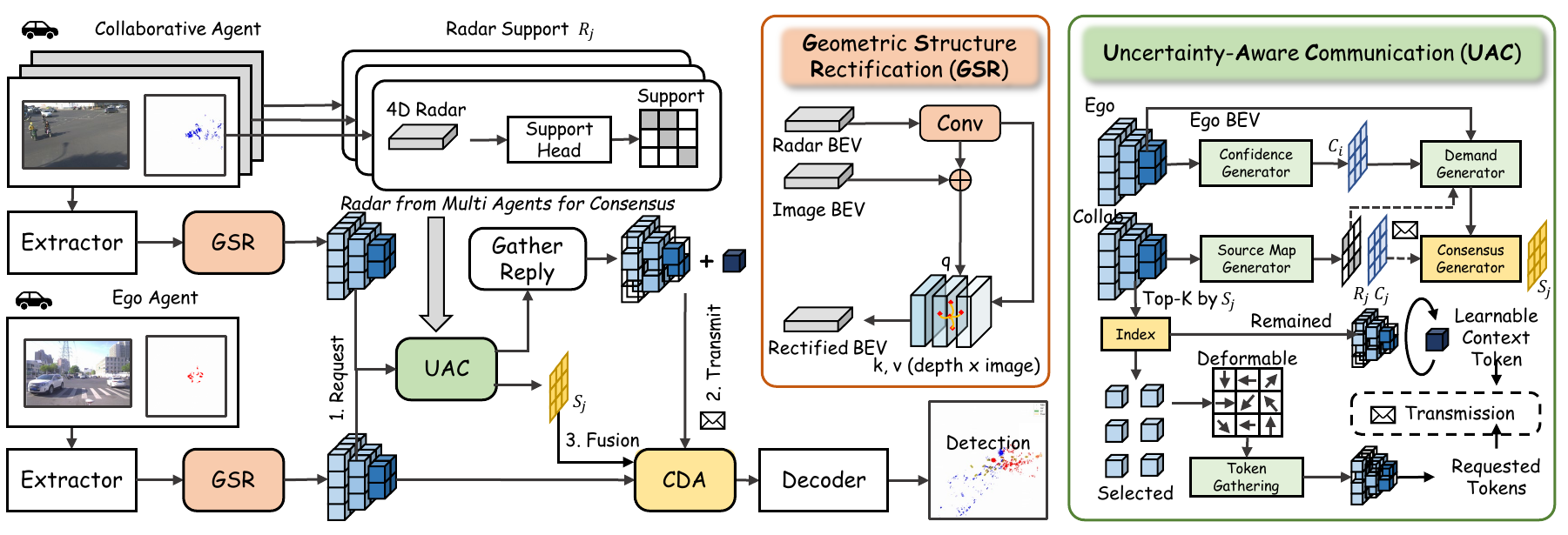}
    \caption{RC-GeoCP framework. GSR uses local radar to ground image features
    in BEV. Each neighbor shares confidence $C_j$ and radar support $R_j$.
    UAC combines them with receiver demand $W_j$ to compute $S_j$ and
    request the top $K$ cells per link and scale. The reply contains
    their feature tokens and one context token for the remaining covered
    cells. AttFuse and CoAlign gather the requested tokens from deformably
    refined features. CDA reuses $S_j$ to weight sources during fusion.}
    \label{fig:framework}
\end{figure*}

\subsection{Selective Communication and Feature Fusion}

Selective communication determines which evidence reaches the receiver,
and fusion determines how much each response contributes to the final
prediction. The value of sharing a region depends on both the information
missing at the receiver and the evidence available from the source. For active
sensing, TITAN--SMART uses spatial and temporal dependencies to choose
where to sense next under a resource budget~\cite{TITANTII2026}.

V2X-ViT fuses LiDAR features~\cite{V2X-ViT}, while CoBEVT fuses camera
features across agents~\cite{CoBEVT}. BM2CP uses local and neighboring
LiDAR returns to constrain image depth before agents share BEV
features~\cite{BM2CP}. CoSDH uses receiver-side LiDAR point density to identify
locations where the receiver needs additional evidence~\cite{CoSDH}. Where2comm guides selection and
fusion with spatial confidence~\cite{Where2comm}. These criteria can
also be applied to radar--camera features, but a gap in sparse radar
coverage leaves object presence uncertain, and high foreground confidence
can survive a visual depth error. RC-GeoCP combines receiver demand with
each source's foreground confidence and learned radar occupancy at the same
cell. The receiver uses this score when it requests a cell and again when
it fuses the features that arrive.

\subsection{Radar Guidance in Collaborative Perception}

DGSM-Net combines Doppler-guided completion with statistical matching
for 4D radar odometry~\cite{DGSMNetTII2026}. In collaborative detection,
sparse returns from each agent provide spatial support for its own
foreground responses.

HRCP first aggregates sources within each modality, then gates the
resulting LiDAR and radar features~\cite{HRCP}. RadarCP uses
Doppler-guided attention in its radar--camera adaptation of BM2CP, which
can downweight stationary targets~\cite{RadarCP}. SlimComm uses motion
and occlusion cues to query neighbors, then applies attention to their
mean response~\cite{SlimComm}. RC-GeoCP keeps radar support separate
for each source and location, so the same evidence guides feature
requests and source weighting after delivery.

\section{Method}

\label{sec:method}
\subsection{Problem and Information Flow}
\label{sec:information_flow}

Receiver $i$ predicts 3D boxes in its own frame using local observations
and a bounded number of features from each neighbor in $\mathcal N(i)$.
Each agent encodes calibrated images $\mathcal I_i$ and 4D radar returns
$\mathcal R_i$ into a multimodal BEV map
$F_i^{(l)}\in\mathbb R^{H_l\times W_l\times c_l}$ at each level $l$, using
radar to guide visual grounding (Sec.~\ref{sec:gsr}).

In Fig.~\ref{fig:framework}, each neighbor first sends two maps of logits,
one for foreground confidence and one for radar support. The receiver
aligns them to its grid by bilinear resampling with the relative pose,
scores each cell and sends the indices of the highest-ranked cells
on each link (Sec.~\ref{sec:uac}). Each neighbor returns feature tokens
at those indices and one context token per scale for its remaining covered
cells. The receiver reuses the score to fuse these replies
(Sec.~\ref{sec:cda}). Spatial supervision and detection feedback train
the score (Sec.~\ref{sec:learning}).

The budget in our evaluation applies to each active link. For receiver $i$,
let $B_{ij}^{\rm map}$, $B_{ij}^{\rm req}$, and $B_{ij}^{\rm reply}$
denote the bytes for source $j$'s control maps, the receiver's request,
and the source's feature and context reply in one update. The receiver's
total payload is
\begin{equation}
    C_i^{\rm total}=\sum_{j\in\mathcal N(i)}
    \bigl(B_{ij}^{\rm map}+B_{ij}^{\rm req}+B_{ij}^{\rm reply}\bigr).
    \label{eq:budgeted_detection}
\end{equation}
Quotas limit requested locations separately for each link and feature
scale (\eqref{eq:selection}). The receiver total therefore varies with
the number of active links and their overlap. The message encoding sets
the number of bytes used for the maps and feature values
(Sec.~\ref{sec:systems}).

Training minimizes detection loss in the receiver frame under this
allocation. At inference, a high-scoring feature can displace another
from the same source. Quotas stay with the assigned neighbor even when
two neighbors cover the same region.

\subsection{Radar-Guided Visual Grounding}
\label{sec:gsr}

Uncertain depth can spread image evidence across BEV cells.
Geometric Structure Rectification (GSR) uses 4D radar to guide frustum
sampling and calibrates the resulting features in BEV.

\subsubsection{4D radar-conditioned image sampling}
At image cell $u$ and depth bin $d$, the depth softmax $P_{i,v}(d,u)$
weights the image feature $I_{i,v}(u)$ to form
$F^{\rm frus}_{i,v}(d,u)=P_{i,v}(d,u)I_{i,v}(u)$. Calibrated lift and voxel
pooling yield the camera map $F_i^{\rm cam}$~\cite{LSS}. With radar map
$F_i^{\rm rad}$, the query at cell $q$ uses twice the camera BEV stride,
\begin{equation}
    Q_i(q)=F_{i,\downarrow}^{\rm cam}(q)+
    \Phi_{\rm init}\bigl(F_{i,\downarrow}^{\rm rad}(q)\bigr)
    +E(q)+E_{\rm pos}(q).
    \label{eq:gsr_query}
\end{equation}
Here $\Phi_{\rm init}$ projects the resized 4D radar features into the query
space, while $E$ and $E_{\rm pos}$ are learned query and positional embeddings.
Calibration projects eight reference heights per BEV cell into normalized
image and depth coordinates $r_{i,v,n}(q)$. The heights are evenly spaced
between the vertical grid bounds inset by 0.5 m. Valid projections lie in
front of the camera and inside the image.

We adapt depth-aware deformable sampling~\cite{SGDet3D} using the
radar-conditioned query $Q_i(q)$. Each of the eight heads predicts offsets
$\delta_{h,n}$ and attention logits $\ell_{h,n}$ for $K=8$ points at one
image level. Softmax over these points gives $a_{h,n}$. Samples outside
the interpolation support contribute zero. The per-view update is
\begin{equation}
    \begin{aligned}
        \xi_{i,v,h,n}(q)&=r_{i,v,n}(q)+\delta_{h,n}(Q_i(q))\oslash(W_I,H_I,D),\\
        Y_{i,v,h}(q)&=\sum_{n=1}^{K}a_{h,n}(q)\,\phi_3(F^{\rm frus}_{i,v,h},\xi_{i,v,h,n}(q)).
    \end{aligned}
    \label{eq:sampling}
\end{equation}
Here $W_I,H_I,D$ are the image feature width, height and number of
depth bins, and $\oslash$ denotes componentwise division. The sampler
$\phi_3$ interpolates depth probabilities at the surrounding image
cells and combines the resulting depth-weighted values bilinearly.
We average the contributing views, apply one deformable self-attention
layer, then upsample and concatenate the result with $F_i^{\rm cam}$ to
obtain $F_i^{\rm rect}$ after projection.

\subsubsection{Gated calibration}
Let $\Psi$ project and resize the radar map to the rectified camera grid.
A learned scalar projection $\mathcal G$ gives the cell-wise gate
$\gamma_i(p)=\sigma(\mathcal G(F_i^{\rm rect})(p))$ for the radar update
\begin{equation}
    \widetilde F_i(p)=F_i^{\rm rect}(p)+\gamma_i(p)\,\Psi(F_i^{\rm rad})(p).
    \label{eq:calibration}
\end{equation}
The gate is broadcast across channels and blends radar features into
each fixed BEV cell.

Sampling and calibration act on different parts of a depth error. If
image evidence has been lifted to the wrong cell, the offsets in
\eqref{eq:sampling} can retrieve image features from another point in the
frustum. Once those samples have been placed on the grid, the gate in
\eqref{eq:calibration} controls how much radar information enters the
feature at that cell. The residual path retains the retrieved image
features even where the radar update is small.

\subsubsection{4D radar feature encoding across viewpoints}
Each agent encodes radar locations and any available motion and intensity
cues. Doppler varies with the line of sight, so each radar head
estimates spatial support from its own encoding. The receiver aligns
these maps to compare sources at the same BEV location.

For a unit line of sight $\hat{\boldsymbol r}_j$ from source $j$ to a
target, the radial component of their relative velocity is
$\hat{\boldsymbol r}_j^\mathsf{T}(\boldsymbol v_{\rm target}-
\boldsymbol v_j)$. Two agents viewing the same moving target from
different bearings can therefore measure different radial velocities.
The radar branch can use motion evidence in the source frame where it
was observed. What the receiver compares after alignment is each
source's predicted occupancy at a common location. This also lets a
return from a stationary object contribute through its position.

\subsection{Receiver-Conditioned Evidence Acquisition}
\label{sec:uac}

Uncertainty-Aware Communication (UAC) prioritizes remote evidence that
can complement the receiver view. It combines local features with source
confidence and radar support to select locations within each quota.

\subsubsection{Control maps and geometric support}
Let $\mathcal S_i=\{i\}\cup\mathcal N(i)$ be the participating source slots.
At level $l$, $v_{i,j}^{(l)}(p)$ marks the cells covered by source $j$ on
the receiver grid. Let $Z_{i,j}^{(l)}$ be the aligned
feature grid after any backbone-specific refinement. We omit the level
superscript where it is clear.

Each source predicts a foreground logit $o_j$ from local multimodal
features and a radar support logit $g_j$ from its 4D radar BEV.
After alignment, $C_{j\to i}=s(\bar o_{ij})$ and
$R_{j\to i}=s(\bar g_{ij})$, where $s$ is a numerically stabilized sigmoid.
$C$ is the source's multimodal foreground confidence. $R$ is a learned
occupancy cue from the radar branch. Its supervision comes from the
aggregate radar prediction in \eqref{eq:collaborative_supervision}.

\begin{figure}[t]
    \centering
    \includegraphics[width=\linewidth]{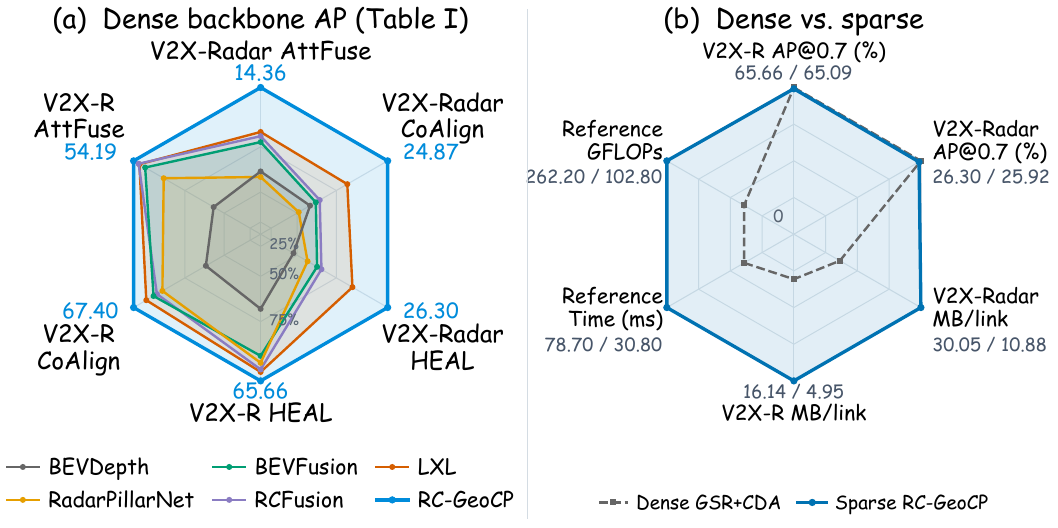}
    \caption{Accuracy and collaboration cost with FP32 exchange.
    (a) AP@0.7 across dense fusion backbones.
    (b) Dense/sparse accuracy, communication, and computation costs.}
    \label{fig:comparison}
\end{figure}

\begin{figure*}[t]
    \centering
    \includegraphics[width=0.95\textwidth]{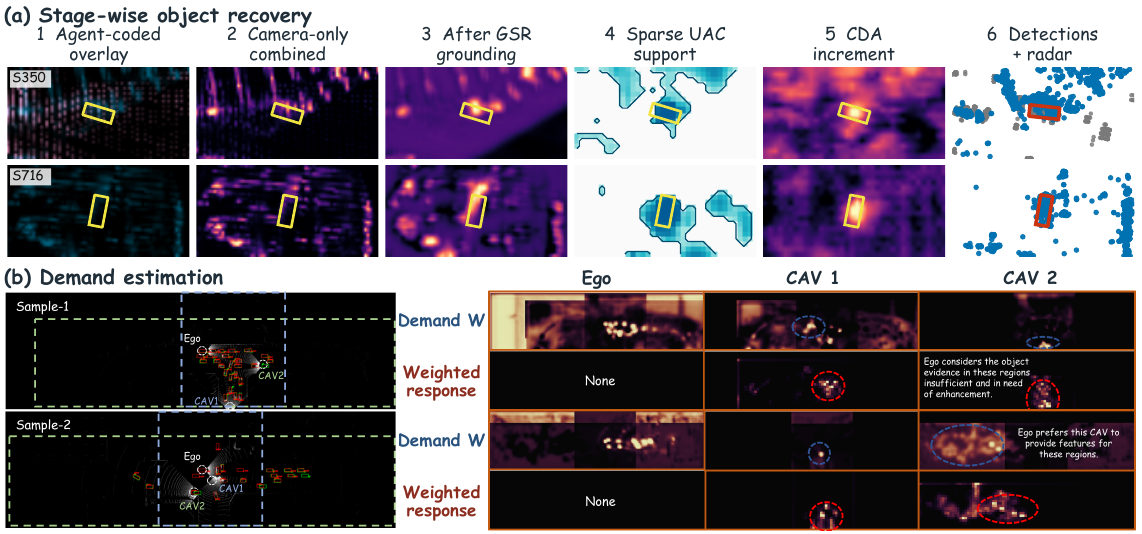}
    \caption{(a) Two object recoveries from camera evidence through 4D radar-guided
    grounding (GSR), sparse acquisition (UAC), and assembly (CDA). Yellow and red boxes
    denote ground truth and detections matched at BEV IoU $\geq 0.70$. Stage heatmaps are
    normalized independently. (b) In each scene, the upper row shows
    demand $W$. The lower row shows the intermediate response weighted
    by the positive source--ego confidence gap. Outlines mark corresponding
    locations. LiDAR returns provide visual context.}
    \label{fig:stages}
\end{figure*}

\subsubsection{Receiver demand and shared scoring}
At each BEV cell, demand is relative preference for a source given
 evidence already available to the receiver. Let $C_i$ be the local
foreground confidence and $C_j,R_j$ the aligned confidence and radar
support from source $j$. We compute the demand weights as
\begin{equation}
    W_{i,j}=\softmax_{j\in\mathcal S_i}
    \bigl(v_{i,j} f_{\rm dem}(Z_{i,i},C_i,1-C_i,C_j,R_j)\bigr).
    \label{eq:receiver_demand}
\end{equation}
The softmax compares sources at each cell. The two convolutional blocks
of $f_{\rm dem}$ compare local and source confidence using local BEV
features, then incorporate radar support. These inputs are available
before feature requests. Combining $W$ with the source
confidence and radar support gives the geometric consensus score $S$
\begin{equation}
    \begin{aligned}
        q_{i,j}(p)&=W_{i,j}(p)C_{j\to i}(p)R_{j\to i}(p),\\
        S_{i,j}(p)&=\frac{q_{i,j}(p)}{\sum_{k\in\mathcal S_i}q_{i,k}(p)+10^{-8}}.
    \end{aligned}
    \label{eq:score}
\end{equation}
At a fixed cell and equal $W$ and $C$, the ratio of two source scores
equals the ratio of their radar supports. A radar multiplier shared by
all sources would cancel in this ratio. Across cells, the denominator
changes. A source can rank highly where all responses are weak if the
other sources have even less support there. Outside source coverage we
pad control logits with $-5$ and demand logits with zero. These finite
values enter both normalizations and the aligned auxiliary predictions.
Only covered cells are eligible for feature requests.

\subsubsection{Quotas and transmitted indices}
Let $\mathcal V_{i,j}^{(l)}=\{p:v_{i,j}^{(l)}(p)=1\}$ and
$n_l=\lfloor\rho_l H_l W_l\rfloor$, with inference fractions
$\rho_l=0.40,0.30,0.20$ from fine to coarse. For nonempty overlap,
$K_{i,j}^{(l)}=\max(1,\min(n_l,|\mathcal V_{i,j}^{(l)}|))$.
Empty overlap yields zero valid tokens (payload accounting in
Sec.~\ref{sec:systems}). We select locations by
\begin{equation}
    \Omega_{i,j}^{(l)}
    =\underset{\substack{\Omega\subseteq\mathcal V_{i,j}^{(l)}\\
                        |\Omega|=K_{i,j}^{(l)}}}{\arg\max}
      \sum_{p\in\Omega} S_{i,j}^{(l)}(p).
    \label{eq:selection}
\end{equation}
Top-$K$ ranks valid cells by the source's relative priority at each
location. The receiver keeps all its local features.

Consider a receiver with one neighbor and a quota of one feature.
At two cells, suppose ego and neighbor have equal $W$ and $C$ within
each cell, and the total unnormalized score is the same at both cells.
The neighbor has the stronger radar response at the first cell and ego
has the stronger response at the second. The neighbor's relative score therefore favors the first cell. The receiver
already has better local support at the second. This advantage depends
on what the radar responses can distinguish. If their relative support
is identical at the two cells, radar supplies no such ordering.

\subsection{Source-Weighted Feature Fusion}
\label{sec:cda}

The Consensus-Driven Assembler (CDA) reuses the acquisition score to
weight sources at each BEV location. For covered locations skipped by
selection, it uses compact source context.

\subsubsection{Receipt masks and source context}
At cell $p$, $\mathcal A_i^{(l)}(p)=\{j\in\mathcal S_i:v_{i,j}^{(l)}(p)=1\}$
contains the sources with valid coverage, including the receiver. The receipt mask is
$m_{i,j}^{(l)}(p)=\mathbf 1[p\in\Omega_{i,j}^{(l)}]$ for neighbors and
$m_{i,i}^{(l)}(p)=1$. Each neighbor sends the requested feature vectors
and one $c_l$-channel context token per level, computed with a learned
query. The receiver projects the token to $e_{i,j}^{(l)}$ and shares this
value across the remaining covered cells
$\mathcal V_{i,j}^{(l)}\setminus\Omega_{i,j}^{(l)}$.

\subsubsection{Location-wise mixture with the acquisition score}
Pyramid CDA maps $S$ to source weights at each cell with $\tau=1$
\begin{equation}
    \alpha_{i,j}(p)=\frac{\exp(S_{i,j}(p)/\tau)}{\sum_{k\in\mathcal A_i(p)}\exp(S_{i,k}(p)/\tau)},\quad j\in\mathcal A_i(p).
    \label{eq:weights}
\end{equation}
Because $0\leq S_{i,j}\leq1$, the ratio of any two covered-source
weights is at most $\exp(1)$ at this temperature. Thus the score can
favor one source while retaining contributions from all covered sources.
For each source, the receipt mask selects its spatial feature or its
context before the weighted sum
\begin{equation}
    \begin{aligned}
        V_{i,j}(p)&=m_{i,j}(p)Z_{i,j}(p)+(1-m_{i,j}(p))e_{i,j},\\
        F_i^{\rm collab}(p)&=\sum_{j\in\mathcal A_i(p)}\alpha_{i,j}(p)V_{i,j}(p).
    \end{aligned}
    \label{eq:assembly}
\end{equation}
Dense exchange uses delivered features across the valid grid. With
only receiver coverage, the direct mixture returns the local feature. HEAL
applies this mixture at every level and resizes, concatenates and
convolves the outputs before detection.

\begin{table*}[t]
    \belowrulesep=0pt
    \aboverulesep=0pt
    \caption{Dense-exchange validation BEV AP by backbone and ego range (m). C and R denote camera and radar.}
    \label{tab:coop_detailed}

    \centering
    \footnotesize
    \setlength{\tabcolsep}{1pt}
    \renewcommand{\arraystretch}{1.00}
    \begin{tabular}{@{}C{33pt}|C{19pt}|C{98pt}|%
        C{24pt}*{2}{C{21.75pt}}|C{24pt}*{2}{C{21.75pt}}|*{4}{C{24.5pt}}|*{4}{C{24.5pt}}@{}}
        \toprule[1.0pt]
        \multirow[c]{3}{*}{Agent} & \multirow[c]{3}{*}{M} & \multirow[c]{3}{*}{Fusion}
        & \multicolumn{6}{c|}{V2X-Radar} & \multicolumn{8}{c}{V2X-R} \\
        \cmidrule(lr){4-9}\cmidrule(lr){10-17}
        & & & \multicolumn{3}{c|}{AP@0.7 $\uparrow$} & \multicolumn{3}{c|}{AP@0.5 $\uparrow$}
        & \multicolumn{4}{c|}{Syn.\ AP@0.7 $\uparrow$} & \multicolumn{4}{c}{Asyn.\ (50~ms) AP@0.7 $\uparrow$} \\
        \cmidrule(lr){4-6}\cmidrule(lr){7-9}\cmidrule(lr){10-13}\cmidrule(lr){14-17}
        & & & Overall & 0--30 & 30--50 & Overall & 0--30 & 30--50
        & Overall & 0--30 & 30--50 & 50--100 & Overall & 0--30 & 30--50 & 50--100 \\
        \midrule[0.5pt]
        \multirow[c]{6}{*}{AttFuse} & C & \mbox{BEVDepth (AAAI'23)~\cite{BEVDepth}} & 8.96 & 16.56 & 3.14 & 29.55 & 50.95 & 9.90 & 31.18 & 43.29 & 26.04 & 23.38 & 22.20 & 36.41 & 16.32 & 11.80\\
        & R & \shortstack{RadarPillarNet (TIM'23)~\cite{RCFusion}} & 8.51 & 14.76 & 3.14 & 25.11 & 38.02 & 15.98 & 46.63 & 66.10 & 44.36 & 28.36 & 41.62 & 59.31 & \apsecond{41.97} & 22.05\\
        & R+C & \mbox{BEVFusion (ICRA'23)~\cite{BEVFusion}} & 11.09 & \apthird{20.41} & \apthird{3.41} & 33.01 & \apthird{55.62} & 16.07 & 51.39 & 70.84 & 51.31 & \apthird{34.76} & 42.77 & 62.77 & 36.47 & \apfirst{27.85}\\
        & R+C & \mbox{RCFusion (TIM'23)~\cite{RCFusion}} & \apthird{11.48} & 20.34 & \apsecond{3.56} & \apthird{33.26} & 54.52 & \apsecond{16.96} & \apsecond{52.99} & \apsecond{72.90} & \apthird{52.16} & \apsecond{34.80} & \apthird{43.68} & \apsecond{64.61} & \apthird{41.94} & 25.08\\
        & R+C & \mbox{LXL (TIV'24)~\cite{LXL}} & \apsecond{11.75} & \apsecond{21.61} & 2.39 & \apsecond{33.46} & \apsecond{55.77} & \apthird{16.57} & \apthird{52.80} & \apthird{72.21} & \apsecond{52.62} & 34.67 & \apsecond{43.75} & \apthird{63.47} & 40.39 & \apsecond{26.35}\\
        & R+C & RC-GeoCP (ours) & \apfirst{14.36} & \apfirst{23.85} & \apfirst{7.63} & \apfirst{35.02} & \apfirst{56.25} & \apfirst{21.91} & \apfirst{54.19} & \apfirst{73.29} & \apfirst{53.11} & \apfirst{35.28} & \apfirst{44.12} & \apfirst{64.99} & \apfirst{42.53} & \apthird{26.01}\\
        \midrule[0.5pt]
        \multirow[c]{6}{*}{CoAlign} & C & \mbox{BEVDepth (AAAI'23)~\cite{BEVDepth}} & 14.72 & 25.93 & 2.55 & 31.77 & 52.83 & 10.11 & 42.24 & 56.40 & 42.43 & 28.74 & 37.87 & 53.18 & 35.36 & 26.05\\
        & R & \shortstack{RadarPillarNet (TIM'23)~\cite{RCFusion}} & 12.74 & 22.32 & 3.64 & 29.26 & 44.28 & 16.96 & 58.43 & 76.83 & 62.64 & 34.98 & 50.23 & 70.17 & 53.13 & 26.25\\
        & R+C & \mbox{BEVFusion (ICRA'23)~\cite{BEVFusion}} & 15.65 & 26.81 & 3.24 & 37.14 & 57.89 & 18.96 & \apthird{61.30} & \apthird{80.35} & \apthird{65.22} & 33.02 & 54.63 & 76.89 & 51.22 & \apsecond{36.01}\\
        & R+C & \mbox{RCFusion (TIM'23)~\cite{RCFusion}} & \apthird{16.26} & \apthird{28.77} & \apthird{4.25} & \apthird{37.60} & \apthird{57.91} & \apsecond{19.90} & 60.22 & 78.69 & 64.08 & \apthird{40.10} & \apsecond{56.07} & \apsecond{79.42} & \apsecond{55.33} & \apthird{35.19}\\
        & R+C & \mbox{LXL (TIV'24)~\cite{LXL}} & \apsecond{20.11} & \apsecond{33.85} & \apsecond{4.45} & \apsecond{40.46} & \apsecond{61.60} & \apthird{19.02} & \apsecond{63.57} & \apsecond{82.63} & \apsecond{66.51} & \apsecond{45.11} & \apthird{54.82} & \apthird{77.82} & \apthird{54.26} & 33.18\\
        & R+C & RC-GeoCP (ours) & \apfirst{24.87} & \apfirst{40.53} & \apfirst{8.75} & \apfirst{44.77} & \apfirst{65.60} & \apfirst{28.68} & \apfirst{67.40} & \apfirst{86.27} & \apfirst{69.51} & \apfirst{48.95} & \apfirst{58.99} & \apfirst{81.91} & \apfirst{57.65} & \apfirst{38.09}\\
        \midrule[0.5pt]
        \multirow[c]{6}{*}{HEAL} & C & \mbox{BEVDepth (AAAI'23)~\cite{BEVDepth}} & 12.41 & 22.94 & 2.21 & 28.78 & 53.39 & 7.36 & 45.13 & 59.09 & 43.77 & 34.45 & 33.94 & 53.74 & 29.03 & 16.65\\
        & R & \shortstack{RadarPillarNet (TIM'23)~\cite{RCFusion}} & 15.13 & 24.69 & 5.93 & 31.61 & 46.87 & 19.32 & 61.05 & 81.70 & 64.21 & 35.19 & 50.11 & 69.62 & \apthird{54.22} & 25.24\\
        & R+C & \mbox{BEVFusion (ICRA'23)~\cite{BEVFusion}} & 16.76 & 27.59 & \apthird{6.68} & 38.61 & \apthird{58.74} & 23.49 & 59.19 & \apsecond{83.94} & 65.97 & 35.65 & \apthird{51.45} & \apthird{77.16} & 50.25 & \apthird{26.96}\\
        & R+C & \mbox{RCFusion (TIM'23)~\cite{RCFusion}} & \apthird{17.47} & \apthird{29.48} & \apsecond{6.82} & \apthird{38.89} & 56.62 & \apsecond{25.48} & \apthird{62.77} & 82.47 & \apthird{68.93} & \apthird{38.25} & \apsecond{57.36} & \apsecond{80.17} & \apsecond{58.51} & \apsecond{32.81}\\
        & R+C & \mbox{LXL (TIV'24)~\cite{LXL}} & \apsecond{21.96} & \apsecond{36.37} & 6.34 & \apsecond{43.17} & \apsecond{64.02} & \apthird{24.92} & \apsecond{63.42} & \apthird{83.29} & \apfirst{70.75} & \apfirst{39.84} & 48.19 & 71.25 & 48.15 & 23.18\\
        & R+C & RC-GeoCP (ours) & \apfirst{26.30} & \apfirst{43.94} & \apfirst{9.27} & \apfirst{46.17} & \apfirst{68.13} & \apfirst{29.30} & \apfirst{65.66} & \apfirst{87.64} & \apsecond{70.15} & \apsecond{39.16} & \apfirst{58.65} & \apfirst{81.87} & \apfirst{60.02} & \apfirst{34.21}\\
        \bottomrule[1.0pt]
    \end{tabular}
\end{table*}

\begin{table}[t]
    \belowrulesep=0pt
    \aboverulesep=0pt
    \caption{BEV detection AP with common radar--camera inputs.}
    \label{tab:benchmark}

    \centering
    \footnotesize
    \setlength{\tabcolsep}{0pt}
    \renewcommand{\arraystretch}{1.00}
    \begin{tabular}{@{}C{76.8pt}|*{2}{C{29pt}}|*{2}{C{29pt}}|*{2}{C{29pt}}@{}}
        \toprule
        \multirow{2}{*}{Method} & \multicolumn{2}{c}{V2X-Radar Val.} & \multicolumn{2}{c}{V2X-Radar Test} & \multicolumn{2}{c}{V2X-R Val.}\\
        \cmidrule(lr){2-3}\cmidrule(lr){4-5}\cmidrule(lr){6-7}
        & @0.5 & @0.7 & @0.5 & @0.7 & @0.5 & @0.7 \\
        \midrule
        Who2com~\cite{Who2com} & 27.10 & 10.35 & 17.51 & 4.28 & 77.73 & 45.57\\
        AttFuse~\cite{OPV2V} & 33.01 & 11.09 & 24.78 & 5.75 & 76.76 & 51.39\\
        CoAlign~\cite{CoAlign} & 37.14 & 15.65 & 29.40 & 12.11 & 77.21 & \apthird{61.30}\\
        CoBEVT~\cite{CoBEVT} & 36.89 & 16.06 & 31.90 & 10.73 & 78.71 & 51.48\\
        HEAL~\cite{HEAL} & 38.61 & 16.76 & 27.80 & 10.10 & \apthird{79.54} & 59.19\\
        Where2comm~\cite{Where2comm} & 40.83 & 17.92 & 33.19 & 11.10 & 75.64 & 44.60\\
        V2X-ViT~\cite{V2X-ViT} & 39.88 & 18.31 & 34.81 & 12.35 & 78.21 & 47.72\\
        CoSDH~\cite{CoSDH} & 41.76 & 22.97 & 36.94 & 15.03 & \apsecond{80.11} & \apsecond{64.00}\\
        SlimComm sel.~\cite{SlimComm} & \apthird{42.25} & \apthird{23.17} & \apthird{39.53} & \apthird{16.40} & --- & ---\\
        RadarCP~\cite{RadarCP} & \apsecond{42.89} & \apsecond{24.22} & \apsecond{40.86} & \apsecond{17.02} & --- & ---\\
        RC-GeoCP (ours) & \apfirst{44.55} & \apfirst{25.92} & \apfirst{42.61} & \apfirst{18.77} & \apfirst{81.90} & \apfirst{65.09}\\
        \bottomrule
    \end{tabular}
    \par\smallskip
    \parbox{\linewidth}{\raggedright
         SlimComm selector uses the published selection rule with GSR and
         CDA. This selector and RadarCP require Doppler, unavailable in V2X-R.}
\end{table}

\subsubsection{Instantiation in attention-based scaffolds}
In AttFuse and CoAlign, the requested indices specify where to gather
tokens from the deformably refined source grid. HEAL gathers tokens
directly from its aligned feature pyramid. In the two attention variants,
the receiver value at each cell
$V_{i,i}$ acts as the query. The filled source values $V_{i,j}$ supply keys
and values, with $m_{i,j}=0$ at padded cells. The content logit is
$a_{i,j}=\langle V_{i,i},V_{i,j}\rangle/\sqrt{c_l}$. The prior is
$r_{i,j}=\max(\alpha_{i,j},10^{-6})$. The resulting attention weights are
\begin{equation}
    \beta_{i,j}(p)=\operatorname{softmax}_{j\in\mathcal S_i}\bigl(a_{i,j}(p)+\log r_{i,j}(p)\bigr).
    \label{eq:attention}
\end{equation}
The attention path averages the filled values with $\beta$. Invalid
slots have $\alpha=0$ and enter attention with the finite prior floor
and their padded or context-filled values. Their attention weights
therefore also depend on the content logits. Each backbone retains its
feature hierarchy and decoder.

In the attention variants the odds between two covered sources factor
into their prior odds and a content term,
$\beta_{i,j}/\beta_{i,k}=(r_{i,j}/r_{i,k})\exp(a_{i,j}-a_{i,k})$.
Consequently, a source with a lower acquisition score can still receive
more weight if its delivered feature agrees more closely with the ego
query. In the direct pyramid mixture, the values enter after the weights
have been set. The same acquisition criterion thus has a different
influence on the final prediction in the two fusion operators.

\subsection{Spatial Supervision and Detection Feedback}
\label{sec:learning}

Spatial supervision trains the signals used to compute the shared score.
Detection feedback further trains the score through feature fusion.

\subsubsection{Anchor-derived targets and aligned supervision}
We derive BEV occupancy targets from detection anchor assignments and
pool them to each feature scale. At level $l$, the local loss directly
supervises the foreground logit $o_j^{(l)}$ in the frame of source $j$.
For cross-source supervision, we use the demand weights $W$ to form two
receiver-frame predictions from the aligned foreground and radar logits,
\begin{equation}
    z_{i,h}^{(l)}=\sum_{j\in\mathcal S_i}W_{i,j}^{(l)}\bar h_{ij}^{(l)},\qquad h\in\{o,g\}.
    \label{eq:collaborative_supervision}
\end{equation}
Both aggregates use the receiver's anchor-derived occupancy target.
Local supervision first provides a reference for each source's foreground
prediction. After alignment, the foreground loss on $z_{i,o}^{(l)}$
trains the demand network to combine those predictions at the receiver.
The radar loss reuses the same demand weights. For a fixed $W$, the
derivative of $z_{i,g}^{(l)}$ with respect to a source radar logit is
that source's demand weight. There is also an indirect path because the
demand network takes radar support as input. The target constrains the weighted sum
$z_{i,g}^{(l)}$. Different source responses can yield the same sum, so
this loss alone cannot determine whether an individual source has placed
an object correctly. The auxiliary focal and Dice losses update the
prediction heads and demand network. Detection supplies further feedback
through the fusion weights.

If all sources agree on both logits, the auxiliary predictions at that
cell are insensitive to the demand weights. Differences in their feature
vectors can still provide a detection gradient through fusion.

\subsubsection{Joint objective and gradients through assembly}
The detection objective combines focal anchor classification, smooth
$\ell_1$ box regression, and orientation classification, weighted by 1, 2 and 0.2, respectively.
The local foreground, aggregated foreground and aggregated radar losses
form $\mathcal L_{\rm aux}^{(l)}$ with weights 1, 0.1 and 0.01. Categorical
focal loss supervises the pixel-depth bins. The total objective is
\begin{equation}
    \mathcal L=\mathcal L_{\rm det}+10\mathcal L_{\rm depth}+\sum_l\omega_l\mathcal L_{\rm aux}^{(l)}.
    \label{eq:loss}
\end{equation}
The three-scale configurations use $(\omega_1,\omega_2,\omega_3)=(0.4,0.2,0.1)$ in
fine-to-coarse order, and the single-scale configuration uses $\omega_1=0.4$.

Top-$K$ membership is detached. At a fixed cell, let
$F=F_i^{\rm collab}$. With the receipt masks and feature values fixed,
the detection gradient through the direct mixture is
\begin{equation}
    \frac{\partial\mathcal L_{\rm det}}{\partial S_j}=\frac{\alpha_j}{\tau}\left\langle\frac{\partial\mathcal L_{\rm det}}{\partial F},V_j-F\right\rangle.
    \label{eq:gradient}
\end{equation}
For a covered source $j$, the derivative depends on how its value
differs from the current mixture in the direction of the detection
gradient. Attention supplies a related path through the log prior
above its floor. Updating the shared score through fusion can change
which cells rank highest on a later forward pass. The discrete request
indices receive no gradient.

\begin{table}[t]
    \belowrulesep=0pt
    \aboverulesep=0pt
    \caption{GSR and CDA ablations on BEVFusion--HEAL (validation AP@0.7).
    Dense and sparse CDA share weights. MB/link includes the messages
    required by each configuration.}
    \label{tab:stages}

    \centering
    \footnotesize
    \setlength{\tabcolsep}{2.5pt}
    \renewcommand{\arraystretch}{1.00}
    \begin{tabular}{@{}C{39pt}|C{28pt}|C{30pt}C{30pt}C{30pt}|C{30pt}|C{30pt}@{}}
        \toprule
        Dataset & Base & Samp. & Calib. & GSR & +CDA & Sparse\\
        \midrule
        V2X-R & 59.19 & 60.11 & 61.02 & 61.99 & \textbf{65.66} & \textbf{65.09}\\
        \cmidrule(lr){2-7}
        MB/link & \multicolumn{4}{c|}{15.77} & 16.14 & 4.951\\
        \midrule
        V2X-Radar & 16.76 & 18.65 & 19.77 & 21.30 & \textbf{26.30} & \textbf{25.92}\\
        \cmidrule(lr){2-7}
        MB/link & \multicolumn{4}{c|}{29.36} & 30.05 & 10.884\\
        \bottomrule
    \end{tabular}
\end{table}
\begin{figure*}[t]
    \centering
    \includegraphics[width=0.90\textwidth]{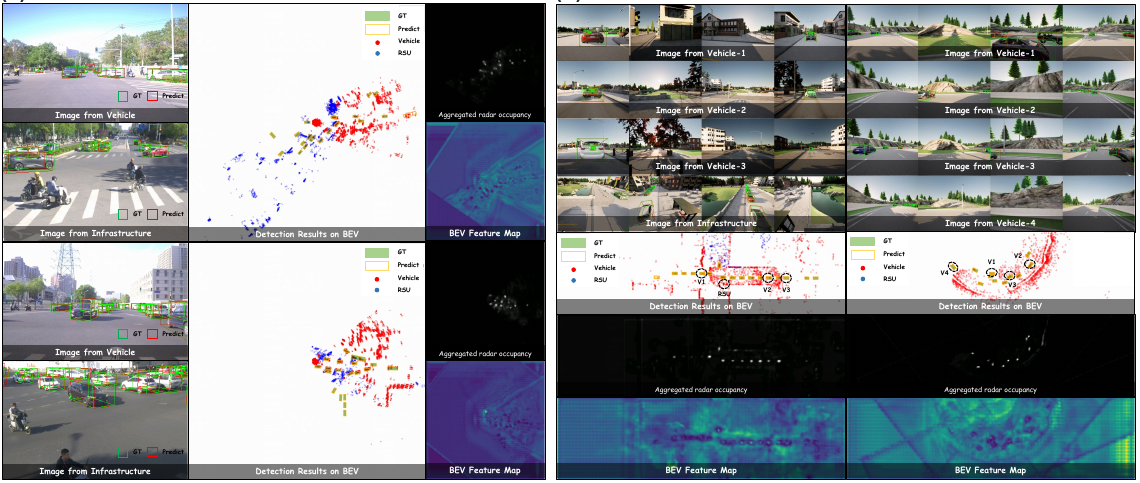}
    \caption{Qualitative results on
    (a) V2X-Radar and (b) V2X-R. Camera views from multiple agents are paired
    with receiver-frame BEV detections and visualizations of the aggregated
    radar prediction $z_{i,g}^{(l)}$ from \eqref{eq:collaborative_supervision}.}
    \label{fig:evidence}
\end{figure*}
\section{Experimental Evaluation}

\label{sec:experiments}

\subsection{Datasets, Training and Evaluation}

We evaluate real-world V2X-Radar~\cite{V2X-RADAR} and simulated
V2X-R~\cite{V2X-R}. V2X-Radar has 30 training, five validation and five
test sequences. Its validation set contains 922 ego--infrastructure
pairs. V2X-R has 8,084 training, 829 validation and 3,166 test frames.
Of its validation ego frames, 552 have two neighbors and 277 have four.
V2X-R uses the camera field of view. The V2X-Radar benchmark evaluator
keeps in-range ground-truth objects whose centers lie in the horizontal field
of view of camera~1 at either the vehicle ($68.40^\circ$ full width) or
the roadside unit ($48.77^\circ$), counting each object ID once per frame.
All in-range predictions surviving nonmaximum suppression enter AP.
For the encoding study, we freeze the V2X-Radar pyramid checkpoint
and fix preprocessing per frame across encodings.
For each encoding, we score its predictions on the same 1,075 test frames against
8,940 targets under the benchmark camera filter and 11,356 under a
radar-coordinate filter. The latter uses vehicle and roadside half-angles
of $50^\circ$ and $56^\circ$, each narrowed by $15^\circ$ beyond 25~m.
We compute BEV AP from the precision--recall envelope at oriented BEV
IoU thresholds of 0.5 and 0.7.

We adapted the baselines in the V2X-Radar repository. Within each
dataset we fix input modalities, splits and sensing region for the
common-input comparison. Unless specified, all methods also share feature
extractors and image resolution. RadarPillarNet~\cite{RCFusion}
encodes radar and EfficientNet~\cite{EfficientNet} encodes images.
Projected LiDAR points supply training targets for 98 uniform depth bins
over 2--100 m. Inference uses cameras and radar.

After depth pretraining, we train for 48 epochs with Adam, batch size
four, weight decay 0.0001 and an initial learning rate of 0.0012 with
cosine decay after 500 warm-up iterations. We evaluate final-epoch
checkpoints. After epoch 12, each scale samples a retention ratio
uniformly from $[0.25,0.75]$ and keeps its top-ranked cells.
Validation sets the default fine-to-coarse quotas at 40/30/20\%.
Experiments use RTX A6000 GPUs.
AP is reported in percent and differences in percentage points.
Paired values below follow V2X-R/V2X-Radar unless specified.

\subsection{Detection across Backbones and Scenes}

With dense exchange, RC-GeoCP leads all three backbone groups in
overall AP@0.7 on both synchronous benchmarks and on V2X-R at 50~ms
delay (Table~\ref{tab:coop_detailed}). Under synchronous HEAL on
V2X-R, it reaches 65.66 overall and leads LXL within 30~m (87.64
vs.\ 83.29). LXL leads in both farther bins, although the 50--100~m
ranking reverses at 50~ms delay.

In the common-input benchmark (Table~\ref{tab:benchmark}), sparse
RC-GeoCP reaches 18.77 AP@0.7 on the held-out V2X-Radar test set,
compared with 17.02 for RadarCP (BM2CP-RCA). On V2X-R validation,
it reaches 65.09 versus 64.00 for the full CoSDH system. The SlimComm
row evaluates its selector with our GSR and CDA.

\subsection{Separating Grounding, Assembly and Sparsification}

On both datasets, the stage ablation uses BEVFusion for local fusion
and HEAL for cross-agent aggregation.

\begin{table}[t]
    \belowrulesep=0pt
    \aboverulesep=0pt

    \caption{Radar-guided sampling and radar input channels (validation AP@0.7).}
    \label{tab:grounding_control}

    \centering
    \footnotesize
    \setlength{\tabcolsep}{0pt}
    \renewcommand{\arraystretch}{1.00}
    \begin{tabular}{@{}C{67.2pt}|C{46pt}C{46pt}|C{46pt}C{46pt}@{}}
        \toprule
        \multicolumn{5}{@{}l@{}}{\emph{(a) Matched-query visual grounding}}\\\hline
        Query & \multicolumn{2}{c|}{V2X-R AP@0.7}
              & \multicolumn{2}{c}{V2X-Radar AP@0.7}\\
        \midrule
        Camera & \multicolumn{2}{c|}{60.32} & \multicolumn{2}{c}{19.37}\\
        4D radar & \multicolumn{2}{c|}{\textbf{61.99}}
                 & \multicolumn{2}{c}{\textbf{21.30}}\\
        \midrule
        \multicolumn{5}{@{}l@{}}{\emph{(b) Radar input channels on V2X-Radar from full model}}\\\hline
        Input & XYZ & +Doppler & +Intensity & +Both\\
        \midrule
        AP@0.7 & 24.22 & 25.08 & 24.84 & \textbf{25.92}\\
        \bottomrule
    \end{tabular}
\end{table}

\begin{figure*}[t]
    \centering
    \includegraphics[width=0.95\textwidth]{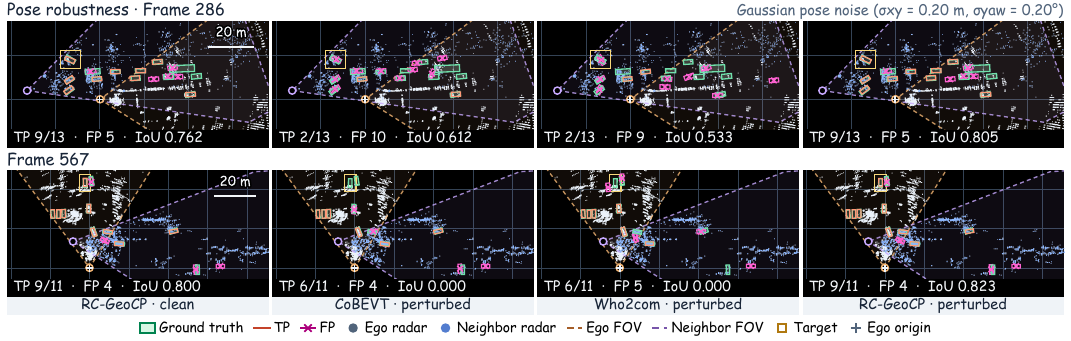}
    \caption{V2X-Radar detections under pose perturbations in Frames 286 (top) and
    567 (bottom). Each row shares ground-truth boxes (green) across methods. Orange
    outlines mark true positives. Magenta outlines with crosses mark false positives.
    Matching is one-to-one in descending confidence order at BEV IoU $\geq 0.70$.
    The reported IoU is the best overlap with the highlighted target in each row.}
    \label{fig:qualitative-comparison}
\end{figure*}

\textbf{Stage contributions.}
Combining sampling and calibration improves dense BEVFusion--HEAL by
2.80/4.54 AP points, exceeding either component alone
(Table~\ref{tab:stages}). With all covered features available, adding
consensus-weighted CDA raises AP by a further 3.67/5.00 points.
Figure~\ref{fig:comparison}(a) reports the dense backbone benchmark.
Panel~(b) switches the stage-ablation model between dense and sparse
exchange at fixed weights. Its dense references are 65.66 on V2X-R and
26.30 on V2X-Radar. At the default quotas, sparse exchange retains
99.1\% and 98.6\% of these AP values while sending 69.3\% and 63.8\%
fewer bytes per link, respectively.

\textbf{Radar-guided visual grounding.}
Adding radar to the image-sampling query improves AP by 1.67/1.93
points with the sampler and calibration fixed
(Table~\ref{tab:grounding_control}(a)). On V2X-Radar, retraining the
full detector with Doppler and intensity alongside XYZ raises AP from
24.22 to 25.92 (panel~(b)). Both local grounding and collaborative
scoring receive the added channels.

\textbf{Qualitative stage progression.}
Figure~\ref{fig:stages}(a) follows two detections from sampling to
fusion. Panel~(b) shows learned demand weights alongside responses
scaled by the positive source--ego confidence gap for display.
Figure~\ref{fig:evidence} shows the aggregated radar prediction
$z_{i,g}^{(l)}$ alongside receiver-frame detections for vehicle--roadside
pairs in V2X-Radar (a) and multiple agents in V2X-R (b).

\subsection{Shared Scoring and Assembly}

\label{sec:shared_analysis}

The score-pairing, fusion and score-sharing comparisons report BEV AP@0.7
on both validation sets. The radar-path ablation uses the V2X-Radar test set.

%
%

\begin{table}[t]
    \belowrulesep=0pt
    \aboverulesep=0pt
    \caption{Consensus and normalized confidence for selection, fusion and neighbor localization.}
    \label{tab:shared_analysis}

    \centering
    \footnotesize
    \setlength{\tabcolsep}{0pt}
    \renewcommand{\arraystretch}{1.00}
    \begin{tabular}{@{}C{51.2pt}|C{50pt}C{50pt}|C{50pt}C{50pt}@{}}
        \toprule
        \multicolumn{5}{@{}l}{\emph{(a) Selection and fusion at fixed weights (validation AP@0.7)}}\\
        \midrule
        \multirow{2}{*}{Acquisition} & \multicolumn{2}{c|}{V2X-Radar} & \multicolumn{2}{c}{V2X-R}\\
        \cmidrule(lr){2-3}\cmidrule(lr){4-5}
        & Assembly $S$ & Assembly $\bar C$ & Assembly $S$ & Assembly $\bar C$\\
        \midrule
        $S$ & \textbf{25.92} & 23.10 & \textbf{65.09} & 61.59\\
        $\bar C$ & 24.88 & 23.34 & 62.19 & 61.52\\
        \midrule
        \multicolumn{5}{@{}l}{\emph{(b) Localization of the selected neighbor (V2X-R validation)}}\\
        \midrule
        \multirow{2}{*}{Ranking score} & \multicolumn{2}{c|}{Pair eligibility} & \multirow{2}{*}{Pairs} & Mean center\\
        \cmidrule(lr){2-3}
        & Min. $C$ & Max. $|\Delta C|$ & & error (m)\\
        \midrule
        Conf. $\bar C$ & \multirow{2}{*}{0.50} & \multirow{2}{*}{0.05} & \multirow{2}{*}{93} & 0.355\\
        Shared $S$ & & & & \textbf{0.186}\\
        \bottomrule
    \end{tabular}

    \caption{Validation AP@0.7 of fusion rules at fixed transmitted locations.}
    \label{tab:assembly_control}

    \centering
    \footnotesize
    \setlength{\tabcolsep}{0pt}
    \renewcommand{\arraystretch}{1.00}
    \begin{tabular}{@{}C{51.2pt}|C{50pt}C{50pt}C{50pt}C{50pt}@{}}
        \toprule
        Dataset & Uniform mean & Sum with $C$ & Sum with $S$ & Full CDA\\
        \midrule
        V2X-R & 61.79 & 62.44 & 64.63 & \textbf{65.09}\\
        V2X-Radar & 21.41 & 22.96 & 24.82 & \textbf{25.92}\\
        \bottomrule
    \end{tabular}

    \caption{Shared vs. independent selection and fusion scores
    (fixed payload, validation AP@0.7).}
    \label{tab:responsibility}

    \centering
    \footnotesize
    \setlength{\tabcolsep}{0pt}
    \renewcommand{\arraystretch}{1.00}
    \begin{tabular}{@{}C{51.2pt}|C{50pt}C{50pt}C{50pt}C{50pt}@{}}
        \toprule
        Dataset & Conf. $C$ & Wider $C$ & Indep. $S$ & Shared $S$\\
        \midrule
        V2X-R & 62.87 & 62.99 & 63.55 & \textbf{65.09}\\
        V2X-Radar & 23.34 & 23.88 & 24.20 & \textbf{25.92}\\
        \bottomrule
    \end{tabular}
\end{table}

\begin{table}[t]
    \centering
    \caption{Radar support in the demand head and score multiplier
    (V2X-Radar test AP@0.7).}
    \label{tab:radar_score_paths}
    \footnotesize
    \begin{tabular}{@{}C{74.4pt}|C{32pt}|C{32pt}|C{32pt}|C{32pt}@{}}
        \toprule
        Demand-head input & $R_j$ & $R_j$ & 0.5 & 0.5\\
        Score multiplier  & $R_j$ & 1.0   & $R_j$ & 1.0\\
        \midrule
        BEV AP@0.7        & \textbf{18.77} & 17.26 & 17.44 & 16.80\\
        \bottomrule
    \end{tabular}
\end{table}

\textbf{Shared scoring and source localization.}
At fixed weights and traffic, using $S$ for both selection and fusion
gains 3.57/2.58 AP points over using source-normalized confidence
$\bar C=C/\sum_j C_j$ for both (Table~\ref{tab:shared_analysis}(a)).
Using $S$ for fusion improves AP under either acquisition rule,
with larger gains when $S$ also selects the cells.
Panel~(b) compares the localization of the neighbor selected by each
score. Across 829 V2X-R validation frames, we run the frozen
CoAlign-based RC-GeoCP independently on each neighbor and match its boxes one-to-one
to ground truth within 4~m in BEV. For two neighbors matched to the same
target, we read both scores from the joint model at the fine-scale cell
midway between their predicted centers, within their common coverage.
The confidence filters retain 93 pairs. Each pair contributes the center
error of the higher-scored box, or the average of both errors for a tied
score. On these high-confidence, near-tied pairs, the full consensus
score reduces mean selected-box error by 16.9~cm relative to $\bar C$.

\begin{figure}[t]
    \centering
    \includegraphics[width=0.95\columnwidth]{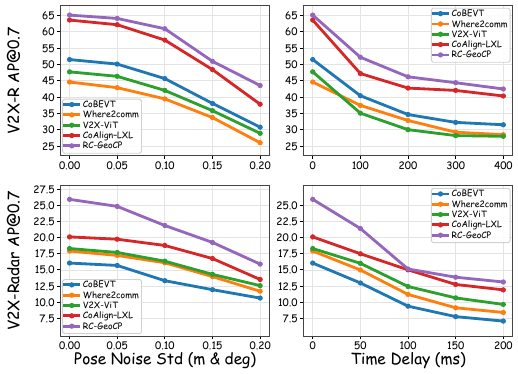}
    \caption{Validation BEV AP@0.7 under pose error and neighbor delay.}
    \label{fig:robustness}
\end{figure}

%

\textbf{Fusion rules.}
At fixed locations and detector weights, replacing $C$ with $S$ in
the same weighted sum gains 2.19/1.86 AP points
(Table~\ref{tab:assembly_control}). The sums pad unused context bytes
to match traffic. Full CDA also uses source context at unrequested
cells and the normalized weights in \eqref{eq:weights}, which together
add 0.46/1.10 points.

\textbf{Shared and independent scores.}
All variants use the same GSR, inputs, quotas and context
(Table~\ref{tab:responsibility}). Wider $C$ widens the confidence head
to approximately match the parameter count of the shared-score model.
Before top-$K$, focal and Dice losses
supervise the auxiliary occupancy predictions of the independent selection
head using receiver-frame targets (Sec.~\ref{sec:learning}). With
top-$K$ detached, detection gradients reach the fusion head. Sharing
that head with selection improves validation AP by 1.54/1.72 points.

\textbf{Radar support after local fusion.}
Each variant in Table~\ref{tab:radar_score_paths} retains local radar--camera
fusion, is retrained separately and uses the same 40/30/20\% evaluation
quotas. With constant inputs in both collaborative paths, V2X-Radar test
AP@0.7 is 16.80. Using $R_j$ in both the receiver demand head and the
score multiplier raises AP to 18.77, a joint gain of 1.97 points.

\subsection{Spatial and Temporal Robustness}

Figure~\ref{fig:qualitative-comparison} shows the effect of pose noise
on individual detections in two V2X-Radar frames.
Figure~\ref{fig:robustness} measures the effects of pose error and
neighbor delay separately. Before alignment, we perturb each agent's
$x$, $y$ and yaw with independent zero-mean Gaussian noise~\cite{RUSS2DTII2026}
of up to 0.20~m and 0.20$^\circ$ standard deviation. Sparse RC-GeoCP
leads CoAlign--LXL on both datasets at the largest perturbation.
Delay tests pair the current receiver frame with older neighbor frames,
up to 400/200~ms, without motion compensation. A moving target can
therefore leave its last radar-supported cell. At 100~ms on V2X-R and
200~ms on V2X-Radar, RC-GeoCP retains leads of 4.99 and 1.22 AP points
over CoAlign--LXL.

\begin{table}[t]
    \belowrulesep=0pt
    \aboverulesep=0pt
    \caption{Payload, fusion cost, and detection AP under sparse and quantized exchange.}
    \label{tab:complexity}

    \centering
    \footnotesize
    \setlength{\tabcolsep}{0pt}
    \renewcommand{\arraystretch}{1.00}
    \begin{tabular}{@{}C{50.4pt}|C{50pt}|C{50pt}|C{50pt}|C{50pt}@{}}
        \toprule
        \multicolumn{5}{@{}l@{}}{\emph{(a) Payload per update (MB)}}\\\hline
        Dataset & Feat. + ctx. & Ctrl. + req. & Per link & Per receiver\\
        \midrule
        V2X-R & 4.525 & 0.426 & 4.951 & 13.209\\
        V2X-Radar & 10.068 & 0.816 & 10.884 & 10.884\\
        \midrule
        \multicolumn{5}{@{}l@{}}{\emph{(b) Reference model size and FP32 collaboration cost}}\\\hline
        Metric & Baseline & LXL-V2VNet & \shortstack{Dense CDA} & RC-GeoCP\\
        \midrule
        Params (M) & 59.8 & 75.4 & 63.0 & 63.8\\
        C-FLOPs (G) & 0.1 & 982.1 & 262.2 & 102.8\\
        C-Lat. (ms) & 1.0 & 165.7 & 78.7 & 30.8\\
        \midrule
        \multicolumn{5}{@{}l@{}}{\emph{(c) Camera-1 and radar horizontal fields of view (AP@0.7)}}\\
        \midrule
        Quota (\%) & Maps / reply & MB/link & Camera~1 & Radar\\
        \midrule
        40/30/20 & FP32 / FP32 & 10.884 & 18.77 & 14.74\\
        \cmidrule(lr){1-5}
        \multirow{3}{*}{7/5/3} & FP32 / FP32 & 2.431 & 18.73 & 14.62\\
        & FP32 / INT8 & 1.142 & 18.69 & 14.58\\
        & UINT8 / INT8 & \textbf{0.626} & 18.65 & 14.55\\
        \bottomrule
    \end{tabular}

\end{table}

\subsection{Communication and Computation}
\label{sec:systems}

The benchmark aligns, selects and fuses features in one forward pass.
A remote source needs the relative pose and receiver grid to reproduce
the aligned descriptor and feature samples. At each scale, the source
sends two control maps, the receiver returns one Int32 index per requested
cell, and the source replies with the selected vectors and a descriptor.
With FP32 maps and replies, $K_l^{\rm msg}$ request slots at level $l$
give the following payload in bytes per link per update
\begin{equation}
    4\sum_l \bigl[(c_l+1)K_l^{\rm msg}+c_l+2H_lW_l\bigr].
    \label{eq:payload_bytes}
\end{equation}
The descriptor and control maps incur a fixed cost on every active
link. With empty overlap the accounting reserves one request slot per
level and the selection mask discards invalid locations. For dense CDA,
the source sends its full feature pyramid and both control maps in one
message. This costs $4\sum_l(c_l+2)H_lW_l$ bytes. In the preceding HEAL
ablations, the receiver can compute foreground confidence from the full
source grids before alignment, so only the feature tensors are transmitted.

\textbf{8-bit message encoding.}
We quantize each source foreground or radar logit map to UINT8 using
its minimum and range before alignment. The selected vectors and
descriptor share a symmetric INT8 scale per channel and level, derived
from their maximum absolute values. The receiver decodes the maps before
scoring and the reply before fusion. With 8-bit maps and replies and
Int32 request indices, the payload in bytes per link per update is
\begin{equation}
    \sum_l \bigl[(c_l+4)K_l^{\rm msg}+5c_l+2H_lW_l+16\bigr].
    \label{eq:quantized_payload_bytes}
\end{equation}
The $4c_l+16$ bytes of decoding parameters per scale are included in
the total.

\begin{table}[t]
    \caption{Validation-selected UAC and CoSDH operating points under
    per-link caps on V2X-Radar (validation/test AP@0.7).}
    \label{tab:review_cosdh_budget}
    \centering
    \footnotesize
    \setlength{\tabcolsep}{0pt}
    \begin{tabular}{@{}C{51.2pt}|C{72pt}C{64pt}C{64pt}@{}}
        \toprule
        Cap (MB) & CoSDH sel. & UAC (ours) & $\Delta$ (pp)\\
        \midrule
        2.50   & 22.14 / 14.42 & \textbf{22.46 / 15.67} & \mbox{$+0.32$/$+1.25$}\\
        5.00   & 23.75 / 16.74 & \textbf{24.61 / 17.45} & \mbox{$+0.86$/$+0.71$}\\
        10.884 & 24.91 / 17.58 & \textbf{25.92 / 18.77} & \mbox{$+1.01$/$+1.19$}\\
        \bottomrule
    \end{tabular}
\end{table}

\textbf{Payload, fusion cost and quantization.}
Table~\ref{tab:complexity}(a) gives the default FP32 payload. A V2X-R
receiver averages 2.67 neighbors and 13.209~MB per update. Panel~(b)
reports reference model sizes and FP32 collaborative neural costs.
The full RC-GeoCP model has 63.8\,M parameters; the dense CDA reference
has 63.0\,M and a different parameterization. The reported RC-GeoCP
FP32 FLOPs and latency are effective estimates that scale with the
transmitted-token fraction. These reference configurations
are distinct from the shared-weight dense/sparse modes in
Table~\ref{tab:stages}.
C-FLOPs count collaborative convolutions and matrix products; C-Lat.
reports collaborative neural latency. Local extraction and network
transport are excluded. Figure~\ref{fig:comparison}(b) uses these same
FP32 reference costs. In Table~\ref{tab:complexity}(c), reducing the
quotas from 40/30/20\% to 7/5/3\% cuts the FP32 payload by 77.7\% to
2.431~MB per link, with 18.73 camera-1 test AP@0.7. Holding these quotas
fixed, 8-bit maps and replies save a further 74.2\% of bytes per link.

\textbf{Computation.}
On an RTX A6000, we profile ten equally spaced frames per test
sequence (50 total), with batch size one, warm-up and five synchronized
repeats per frame. From ready BEV pyramids through collaboration,
pyramid decoding, detection and box postprocessing, median frame-median
times at 7/5/3\% are 30.16~ms for FP32 and 32.02~ms for UINT8/INT8.
Median codec GPU time is 0.66~ms.

\textbf{Per-frame delivery replay.}
We replay every frame at the original cadence. At 100~Mbit/s,
serializing the 0.626~MB payload takes 50.09~ms. A modeled 10~ms base
round-trip time adds 5~ms for each of the three messages, giving
65.09~ms of communication per update.
Using scan-end timestamps as a proxy for local BEV readiness and measured
per-frame GPU time (median 30.85~ms), the replay yields a median
ready-to-output latency of 95.94~ms and a 95th percentile of 101.81~ms.
All 1,075 frames meet the 250~ms deadline. In the radar field of view,
this operating point yields 14.55 AP versus 4.13 for ego alone.

%
%
%

\textbf{Selection under a link cap.}
The 7/5/3\% quotas were fixed separately for the encoding study.
For the selector comparison in Table~\ref{tab:review_cosdh_budget}, we hold
GSR and CDA fixed and choose each selector's operating point on validation
under each per-link cap. At 2.50~MB, UAC first selects cells at
40/30/20\% across scales, then truncates the selection to fit the byte
budget. It reaches 15.67 test AP, 1.25 points above CoSDH. The gain at
5.00~MB is 0.71 points. With the full 40/30/20\% selection, both
selectors send 10.884~MB per link and UAC reaches 18.77 test AP,
1.19 points above CoSDH.

\section{Conclusion}

In this work, we present RC-GeoCP, a geometric consensus framework for
4D radar--camera collaborative perception. Specifically, we propose Geometric
Structure Rectification (GSR) to leverage local radar to ground visual features
in BEV. Then, Uncertainty-Aware Communication (UAC) uses radar support
from each neighbor to combine source confidence with receiver demand and
rank cells for transmission. After delivery, the Consensus-Driven
Assembler (CDA) reuses that score to weight each source during fusion.
Experiments span two radar-based collaborative benchmarks and three backbones. At fixed traffic,
shared scoring yields the highest validation AP among four selection--fusion
pairings on both datasets. Quantized sparse exchange also meets the
modeled deadline on every frame in the V2X-Radar replay. Future work
will extend our framework with adaptive bandwidth allocation.

\bibliographystyle{IEEEtran}
\bibliography{references}
\end{document}